\documentclass[final]{SGSNet}

\usepackage{times}
\usepackage{epsfig}
\usepackage{graphicx}
\usepackage{amsmath}
\usepackage{amssymb}

\usepackage{algorithm}
\usepackage{algorithmicx}
\usepackage{algpseudocode}

\usepackage{color} 
\floatname{}{algorithm}

\usepackage{mathtools}

\usepackage{multirow}
\usepackage{tabu}

\usepackage[pagebackref=true,breaklinks=true,colorlinks,bookmarks=false]{hyperref}

\begin{document}

\title{ 
Semi-Global Shape-aware Network
}

\author{Pengju Zhang$ ^{1,2} $ \ \ \ \ \ \   Yihong Wu$ ^{1,2,}\thanks{Yihong Wu (yhwu@nlpr.ia.ac.cn) is the corresponding author.}$ \ \ \ \ \ \   Jiagang Zhu$ ^{3} $\\
$ ^1 $ School of Artificial Intelligence, University of Chinese Academy of Sciences\\
$ ^2 $ National Laboratory of Pattern Recognition, Institute of Automation, Chinese Academy of Sciences\\
$ ^3 $XForwardAI Technology Co.,Ltd.
}

\maketitle

\begin{abstract}
Non-local operations are usually used to capture long-range dependencies via aggregating global context to each position recently. However, most of the methods cannot preserve object shapes since they only focus on feature similarity but ignore proximity between central and other positions for capturing long-range dependencies, while shape-awareness is beneficial to many computer vision tasks. In this paper, we propose a Semi-Global Shape-aware Network (SGSNet) considering both feature similarity and proximity for preserving object shapes when modeling long-range dependencies. A hierarchical way is taken to aggregate global context. In the first level, each position in the whole feature map only aggregates contextual information in vertical and horizontal directions according to both similarity and proximity. And then the result is input into the second level to do the same operations. By this hierarchical way, each central position gains supports from all other positions, and the combination of similarity and proximity makes each position gain supports mostly from the same semantic object. Moreover, we also propose a linear time algorithm for the aggregation of contextual information, where each of rows and columns in the feature map is treated as a binary tree to reduce similarity computation cost. Experiments on semantic segmentation and image retrieval show that adding SGSNet to existing networks gains solid improvements on both accuracy and efficiency. 	
\end{abstract}

\section{Introduction}
Deep learning, based on local convolutions, has achieved great success in many fields. Recently, many studies show that modeling long-range dependencies can attain consistent improvements in many computer vision tasks, such as semantic segmentation \cite{huang2019ccnet, song2019learnable}, object detection \cite{wang2018non}, and image retrieval \cite{ng2020solar}. Long-range dependencies, containing global contextual information, are usually captured by using local convolutional operations \cite{he2016deep,chen2017deeplab} or non-local blocks \cite{huang2019ccnet, song2019learnable,wang2018non}. 

Conventional convolutions operate data in a local window and many methods enlarge receptive fields by repeating local convolutional operations for capturing long-range dependencies \cite{simonyan2014very, szegedy2015going}. These operations are inefficient to model long-range dependencies and may encounter optimization difficulties \cite{he2016deep}. Considering the problems, several non-local operations are provided to model long-range dependencies, such as Non-local networks \cite{wang2018non}, CCNet \cite{huang2019ccnet}, and GCNet \cite{cao2019gcnet}. In these methods, each spatial position in a feature map is seen as a node, and each node can get supports from all other nodes according to a weight function. The weight function is usually a dot-product similarity between central and other nodes. This way just considers similarity but ignores proximity between two nodes when capturing long-range dependencies. Due to the absence of spatial distances when capturing long-range dependencies, object shapes cannot be modeled. Actually, shape-awareness has been confirmed to be useful in many computer vision tasks \cite{dai2017deformable,zhu2019deformable,luo2020aslfeat}. Tree filters \cite{yang2014stereo,song2019learnable} are adopted to preserve object structures, where minimum spanning trees in low-level feature maps are established and then the process of global context aggregation on high-level feature maps is performed. The efficiencies of spanning trees based methods can still be developed furthermore. 

\begin{figure}
	\centering
	\centerline{\includegraphics[width=8.0cm]{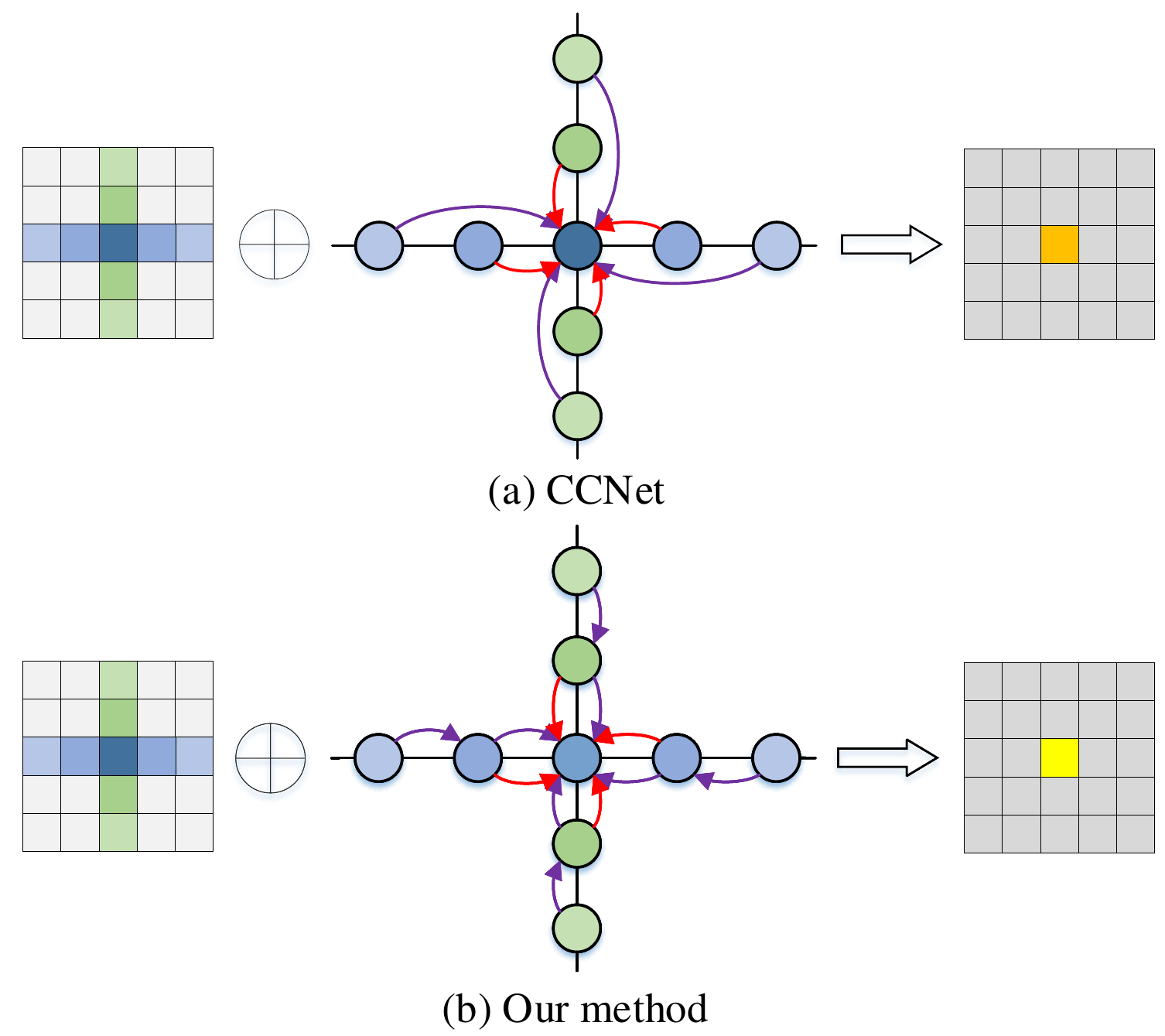}}
	\caption{Differences between our method and CCNet \cite{huang2019ccnet}. "$ \curvearrowright $" means the weight between two nodes. The weights between the central node and the nearest neighbors depend on "$ \curvearrowright $" in red. The weights between the central node and the second nearest neighbors depend on "$ \curvearrowright $" in purple. (Best viewed in color) } 
	\label{comparison}
	\vspace{-1.1em}
\end{figure}

In this paper, we propose a Semi-Global Shape-aware Network (SGSNet) to address the above-mentioned problems. 
In order to consider both feature similarity and proximity for preserving object shapes when modeling long-range dependencies, we aggregate global contextual information for each position in a hierarchical way. In the first level, each position in the whole feature map only aggregates contextual information in vertical and horizontal directions. Then the result of previous level is input into the next level to do the same operations. By this hierarchical way, each central position gains supports from all other positions, and the combination of similarity and proximity makes each position gain supports mostly from the same semantic object. We also propose an efficient algorithm reducing the computational complexity $ \mathcal{O}(N\sqrt{N}) $ of the brute force implementation to $ \mathcal{O}(N) $. 
As a contrast, the computational complexities of Non-local Neural Networks \cite{wang2018non} and CCNet \cite{huang2019ccnet} are $ \mathcal{O}(N^{2}) $  and $ \mathcal{O}(N\sqrt{N}) $ respectively. The computational complexity of \cite{song2019learnable} is also $ \mathcal{O}(N) $ while it needs extra time and computations to establish minimum spanning trees.

Differences between our method and CCNet are illustrated in Fig. \ref{comparison}.
The weight of two nodes just depends on similarity between themselves for CCNet. In our method, the weight of two nodes considers both similarity and proximity for preserving object shapes when capturing long-range dependencies.
The details are described in Sec. \ref{SGSNet}. 

In summary, our main contributions are:
\begin{itemize}
	\item We propose a SGSNet which considers both feature similarity and geometric proximity for preserving object shapes when modeling long-range dependencies.
	\item For practical usages, we propose an efficient algorithm for the aggregation of contextual information which reduces the computational complexity $ \mathcal{O}(N\sqrt{N}) $ to $ \mathcal{O}(N) $. Thus SGSNet can be plugged into existing convolutional neural networks conveniently. 
	\item Experiments on different computer vision tasks (semantic segmentation and image retrieval) show that adding SGSNet to existing deep neural networks achieves higher accuracies while takes less computations. 
\end{itemize}

\section{Related Work}

\subsection{Self-attention}
Self-attention \cite{vaswani2017attention} was initially applied in machine translation area. Non-local Networks \cite{wang2018non} bridged self-attention modules in machine translation to non-local filtering operations in computer vision. 
Many methods improved weight functions in self-attention module to learn discriminative feature representation. Non-local Networks \cite{wang2018non} discussed four choices of weight functions, i.e. Gaussian, Embedded Gaussian, Dot product and Concatenation. Considering the computational complexity $ \mathcal{O}(N^{2}) $ of Non-local Networks, CCNet \cite{huang2019ccnet} adopted recurrent sparse criss-cross attention modules to substitute the dense attention in Non-local Networks. By two consecutive criss-cross attention modules, every node could collect contextual information from all nodes in feature maps. This way reduced the computational complexity $ \mathcal{O}(N^{2}) $ for Non-local Networks to $ \mathcal{O}(N\sqrt{N}) $, therefore it was more efficient, while object shapes could be further considered. 
Song et al. \cite{song2019learnable} first built minimum spanning trees (MST) in low-level feature maps and then computed feature similarity in high-level semantics to preserve object structures. However extra time and computations to build the minimum spanning trees were needed. In this work, we consider both spatial distances and feature similarity when capturing long-range dependencies for higher accuracies meanwhile maintaining a low computational complexity.

\subsection{Semantic Segmentation}
Semantic segmentation is an essential and challenging task in computer vision community. The methods based on Convolution Neural Networks (CNNs) have made significant achievements in the past few years. Long et al. \cite{long2015fully} applied fully convolutional networks (FCNs) in semantic segmentation. Later, researchers found that FCNs were limited by receptive fields due to fixed geometric structures. 
The works of U-net \cite{ronneberger2015u} and DeepLabv3+ \cite{chen2018encoder} used encoder-decoder structures to combine high-level and low-level features for semantic segmentation tasks. Chen et al. \cite{chen2017deeplab} adopted atrous convolutions which could effectively enlarge receptive fields when aggregating contextual information. Furthermore, they also proposed atrous spatial pyramid pooling (ASPP) to complete the segmentation task at multiple scales. Zhao et al. \cite{zhao2017pyramid} exploited global context information by pyramid pooling modules to achieve better performance. PSANet \cite{zhao2018psanet} relaxed the local neighborhood constraint through a self-adaptively learned attention mask. GCN \cite{peng2017large} found that large convolutional kernels were also important when performing a dense per-pixel prediction task and proposed Global Convolutional Network (GCN). Unlike the previous methods, we aggregate global context information in a self-attention manner. 

\subsection{Image Retrieval}
Traditionally, bag-of-visual-words \cite{philbin2007object}, VLAD \cite{arandjelovic2013all} and Fisher vector \cite{perronnin2010large} are usually used to aggregate a set of handcrafted local features \cite{lowe2004distinctive} into a single global vector to represent an image. 
Recently, many methods attempt to replace handcrafted ones by learned counterparts and then aggregate learned features with these similar techniques as the traditional ones \cite{arandjelovic2016netvlad, liu2019stochastic, ge2020self}. Some studies show that directly using a pooling operation \cite{radenovic2018fine} to substitute the aggregation process can get comparable performances. We follow \cite{ng2020solar} to name these methods with a pooling operation as \textit{global single-pass} methods for they do not separate extraction and aggregation steps explicitly. 
Radenovi{\'c} et al. \cite{radenovic2018fine} proposed GeM pooling which generalized average and max pooling and got excellent results.
Based on GeM, SOLAR \cite{ng2020solar} employed second-order similarity and attention for image retrieval and obtained significant performance improvements. In this paper, we also explore the proposed SGSNet components for image retrieval. 

\section{Semi-Global Shape-aware Network}
\label{SGSNet}
In this section, we first introduce preliminary formulations which consider feature similarity and geometric proximity for preserving object shapes when modeling long-range dependencies. 
Then we present the proposed semi-global shape-aware Network. Finally, we propose a linear time algorithm for implementing the network.  

\subsection{Preliminary Formulations}
\label{Formulation}
First, a given feature map $ I $ in CNNs can be seen as a connected, undirected graph $ G = (N, E) $. The nodes $ N $ are all spatial positions in $ I $ and the edges $ E $ with weights $ \omega $ are all connections between two neighbor nodes. 
Next, we define a weight function between different nodes. Let us consider a simple case: given a pair of neighbor nodes $ u $ and $ v $, the weight between $ u $ and $ v $ is defined as
\begin{equation}\label{neighborweight}
\omega(u, v) = \omega(v, u) = exp(-d(u, v)),
\end{equation}
where $ d(u, v) $ is Euclidean distance between feature vectors of $ u $ and $ v $. When nodes $ u $ and $ v $ are not neighbors, the weight function is given by
\begin{equation}
\label{notneighborweight}%
\Omega(u, v)=\prod_{\mathclap{(v_i,v_j)\in P_{u,v}}}\omega(v_i,v_j)=exp(-\sum_{(v_i,v_j)\in P_{u,v}} d(v_i,v_j)),
\end{equation}
where $ v_i $ and $ v_j $ are neighbor nodes in the shortest path $ P $ between $ u $ and $ v $. We denote the feature map after aggregating contextual information as $ I' $. The aggregation function which takes feature similarity and geometric proximity into account simultaneously is:
\begin{equation}\label{aggr}
I'_u=\dfrac{1}{S_u}\sum_{\forall v \in \mathcal{R}}\Omega(u, v)f(I_v) + I_u,
\end{equation}
where $ I'_u $ and $ I_u $ denotes the feature vector at $ u $ in  $ I' $ and $ I $ respectively,
$ \mathcal{R} $ is all nodes in the graph $ G $, the function $ f(\cdot) $ means feature transformation and the weight $ \Omega(u, v) $ is normalized by $ S_u = \sum_{\forall v \in \mathcal{R}}\Omega(u, v) $.

In the above aggregation process, a problem is that there may be a lot of paths between arbitrary nodes $ u $ and $ v $ and it is time-consuming to find the shortest path for all pairs of nodes. In \cite{yang2014stereo, song2019learnable}, the authors adopt a spanning tree to remove "unwanted" edges in the four-connected planner graph $ G $, thus all the nodes are connected by a minimum spanning tree. The minimum spanning tree can enlarge the geometric proximity between two neighbor nodes if these two nodes are dissimilar in appearance. As a result, low support weights will be assigned between these two nodes which causes less robustness to textures \cite{yang2014stereo}. Besides, establishing and traversing a minimum spanning tree needs to pay extra time and computations. Inspired by \cite{hirschmuller2007stereo, huang2019ccnet}, we adopt a semi-global manner to overcome this difficulty by considering the supports from nodes in horizontal and vertical directions in a single block, and then add multiple blocks hierarchically for capturing full-image dependencies. The details are described in Sec. \ref{SemiGlobalBlocks}.
\begin{figure}
	\centering
	\centerline{\includegraphics[width=8.0cm]{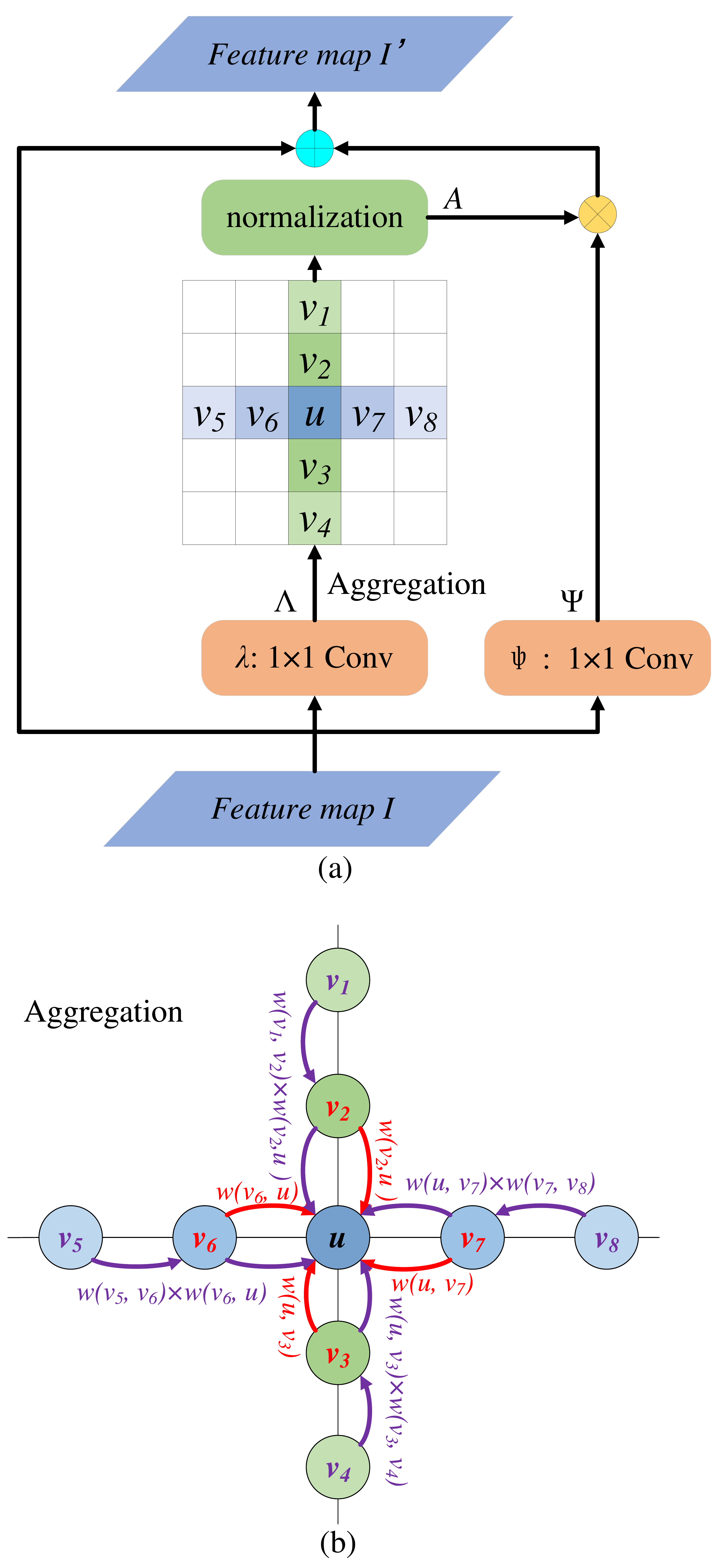}}
	\caption{(a) Architectures of Semi-Global Blocks; (b) illustration of the detailed aggregation, where "$ \otimes $" means multiplication and "$ \oplus $" means element-wise sum. (Best viewed in color)}
	\label{sgsblock}
\end{figure}

\subsection{A Semi-Global Block}
\label{SemiGlobalBlocks}
Due to the difficulty of minimizing matching costs in a 2D image, SGM \cite{hirschmuller2007stereo} utilizes a semi-global approach to aggregate the matching costs along multiple 1D directions. CCNet \cite{huang2019ccnet} just collects contextual information in a criss-cross path. In this paper, we establish a semi-global block, as shown in Fig. \ref{sgsblock} (a). Given an input feature map $ I \in \mathcal{R}^{C \times W \times H} $ with channels $ C $, width $ W $ and height $ H $, the block first utilizes a 1 $ \times $ 1 convolution $ \lambda $ on $ I $ to reduce $ C $. The resulting feature map is denoted as $ \Lambda \in \mathcal{R}^{C' \times W \times H} $ with $ C' < C $. We compute an attention map $ A \in \mathcal{R}^{(H+W-1) \times W \times H} $ by a weight function on $ \Lambda $. Channels at a position of $ A $ correspond to the weights between this position and other positions in the same column or row respectively. The block also applies another 1 $ \times $ 1 convolution $ \psi $ on $ I $ and the resulting feature map $ \Psi $ has the same size as $ I $. For each position $ u $ in the spatial dimension of $ \Psi $, we can obtain a feature vector $ \Psi_u \in \mathcal{R}^C $. We collect the feature vectors of all positions in horizontal and vertical directions of $u$ on $ \Psi $, resulting in a matrix $ \Theta_u \in \mathcal{R}^{(H+W-1) \times C} $. We denote the output feature map of the block as $ I' \in \mathcal{R}^{C \times W \times H} $, and the feature vector at position $ u $ on $ I' $ by (\ref{aggr}) is:
\begin{equation}\label{aggrblock}
I'_u = \sum_{d \in (0,H+W-1)}A_{u_d}\Theta_{u_d} + I_u,
\end{equation}
where $ d $ is the index of channels of $ A $, $ A_{u_d} $ denotes the value at the $ d^{th} $ channel in position $ u $ on $ A $. $ \Theta_{u_d} $ denotes the $ d^{th} $ feature vector in $ \Theta_u $, and $ I_u $ denotes the feature vector at position $ u $ on $ I $.

It is worth noting that the weight function is quite different from \cite{huang2019ccnet}. The aggregation process is shown in Fig. \ref{sgsblock} (b), where the weight between nodes $ u $ and $ v $ depends on all nodes in the path connecting $ u $ and $ v $. The path is limited in horizontal and vertical directions. Actually, our weight function (\ref{notneighborweight}) in the aggregation process considers feature similarity and geometric proximity simultaneously, as described in Sec \ref{Formulation}.
Therefore, our method can preserve object shapes more clearly compared with \cite{huang2019ccnet}, which will be shown in Sec. \ref{visualization}. 
In order to adjust feature similarity between nodes $ u $ and $ v $, we further add learnable parameters $ \alpha $ and $ \beta $ in (\ref{notneighborweight}):
\begin{equation}\label{weighfunction_p}
\left\{
	\begin{array}{lr}
		\Omega(u, v) 
		 = exp(-\dfrac{D(u,v)}{\alpha}) \ \ \ for \ horizontal &\\
		\Omega(u, v) 
		 = exp(-\dfrac{D(u,v)}{\beta})  \ \ \ for \  vertical,
	\end{array} 
\right.
\end{equation}
where $ D(u,v) = \sum_{(v_i, v_j) \in P_{u,v}} d(v_i, v_j) $,  $ v_i $ and $ v_j $ are neighbor nodes in horizontal or vertical path $ P $ connecting $ u $ and $ v $. Considering that width and height of feature maps may be different, we use $ \alpha $ and $ \beta $ for horizontal and vertical directions respectively. If $ \alpha $ ($ \beta $) is smaller, feature similarity plays a more important role in (\ref{weighfunction_p}), and vice versa. Therefore, the learnable parameter $ \alpha $ ($ \beta $) can adjust the relation of feature similarity and geometric proximity adaptively.

\subsection{Hierarchical Semi-Global Blocks}
\begin{figure}
	\centering
	\centerline{\includegraphics[width=8.0cm]{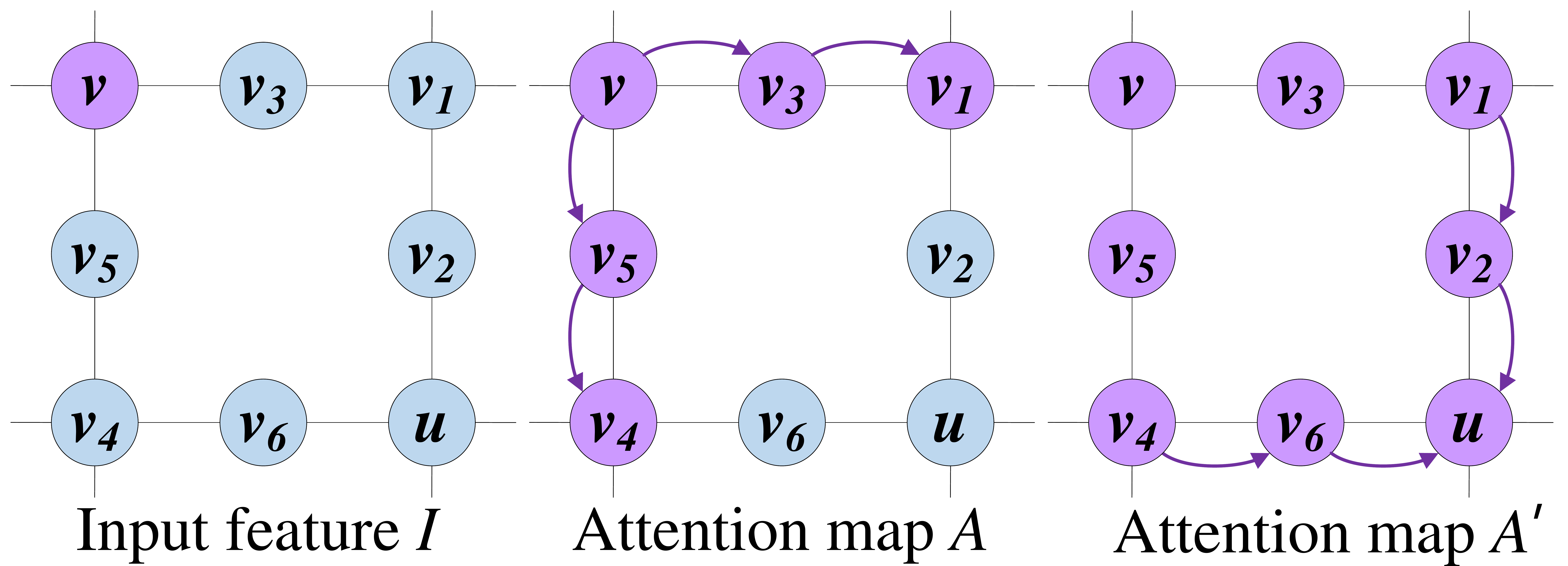}}
	\caption{ Illustration of the process of  spreading contextual information in node $ v $ (in purple) to node $ u $ which is not in the same row and column as $ v $. The node in purple means that this node contains contextual information of node $ v $. The left is an input feature $ I $; the middle is the attention map $ A $ in the first semi-global block; the right is the attention map $ A' $ in the second semi-global block. (Best viewed in color)}
	\label{propagation}
\end{figure}

A single semi-global block can only aggregate contextual information in the same row and column of a central node but ignores other directions. We address this problem by multiple hierarchical semi-global blocks. In the first hierarchical level, a feature map $ I $ is input into a semi-global block, producing a feature map $ I' $. Then in the second hierarchical level, another semi-global block takes $ I' $ as the input and outputs a feature map $ I'' $. Thus, every node in $ I'' $ can aggregate contextual information from all nodes in $ I $. Specifically, we denote the attention maps in the first and second hierarchical levels as $ A $ and $ A' $. The process of spreading contextual information in node $ v $ (in purple) to node $ u $ which is not in the same row and column as $ v $ is illustrated in Fig. \ref{propagation}. In the first semi-global block, only nodes in the same row and column can receive the contextual information from $ v $, which can be seen at $ A $ in Fig. \ref{propagation}. In the second semi-global block, $ u $ receives contextual information from $ v_1 $ or $ v_4 $ which have aggregated contextual information from $ v $ in the first semi-global block, as shown at $ A' $ in Fig. \ref{propagation}. Thus, $ u $ obtains contextual information of $ v $ through the above process.

More generally, every node can capture full-image contextual information by the hierarchical semi-global blocks, which can enhance the capability of a single semi-global block for modeling long-range dependencies \cite{huang2019ccnet}.

\subsection{A Linear Time Algorithm for Efficient Implementation}
\label{LinearTime}
For a clearer explanation, we ignore 1 $ \times $ 1 convolution $ \lambda $ and $ \psi $ in Fig. \ref{sgsblock}, which does not influence our conclusion. In a semi-global block, one node is supposed to aggregate contextual information of those in vertical and horizontal directions. 
According to (\ref{aggrblock}), the total computational complexity of a single semi-global block is $ \mathcal{O}(N\sqrt{N}) $ by the brute force implementation. Specifically, given an input feature map $ I $ with $ N=W \times H $ nodes, every node needs to aggregate contextual information from other nodes in the same row and column. 
Thus, we need $ (H+W-1) $ computation times of similarity for every node. The total computation is $ \mathcal{O}((H \times W) \times (H+W-1)) $, which is unfavorable for practical applications.

\begin{figure}
	\centering
	\centerline{\includegraphics[width=8.0cm]{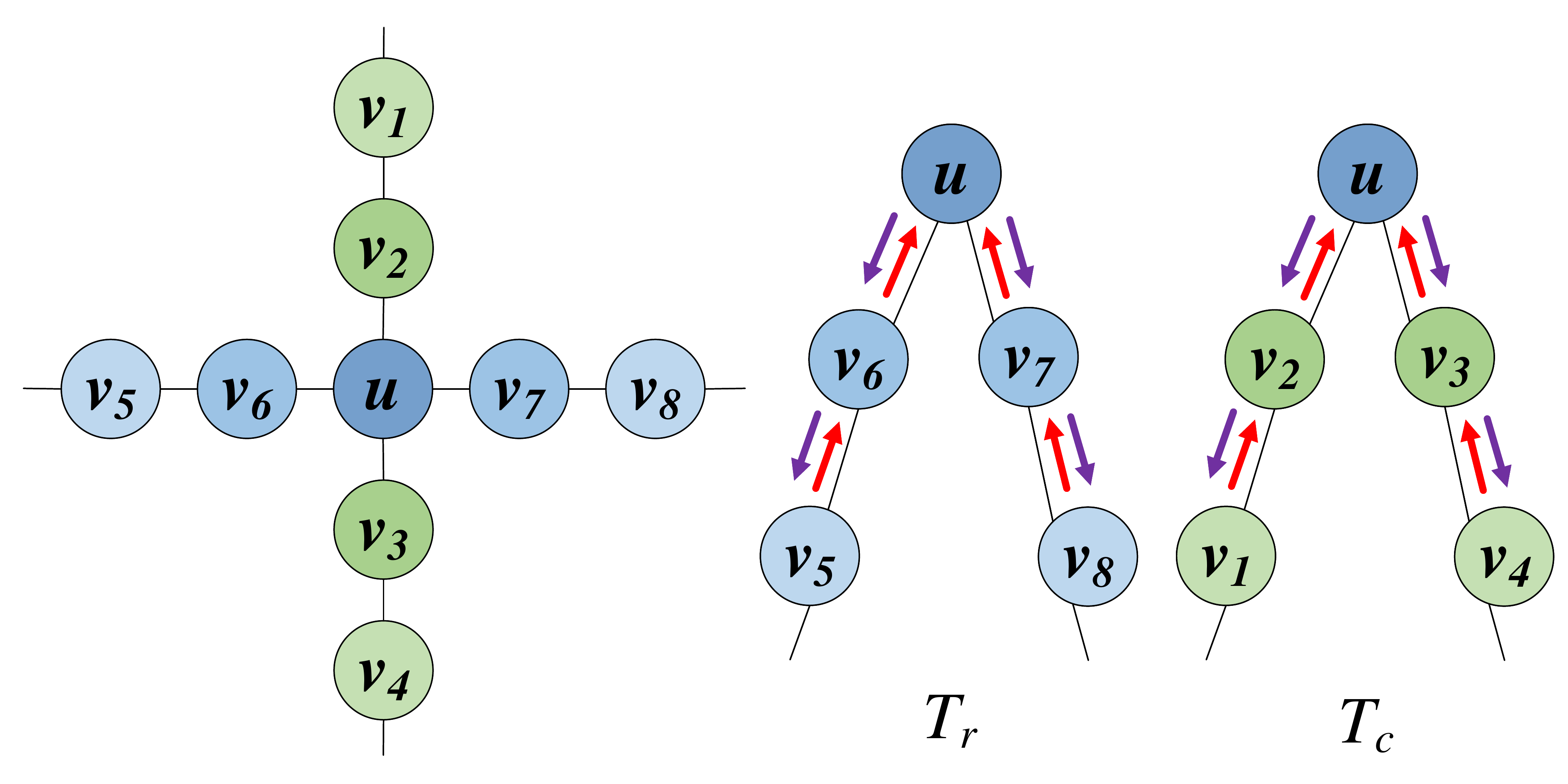}}
	\caption{ Illustration of aggregation and updating steps. The left is a central node $ u $ along with nodes in the same row and column. The right two are binary trees ($ T_r, T_c $) corresponding to the row and column in the left. "$ \rightarrow $" in red and purple mean the aggregation and updating step respectively. (Best viewed in color)}
	\label{aggrProp}
\end{figure}

Noticing that there are extensive repeated computations in the brute force implementation, we propose an optimized alternative which reduces the above computational complexity of a single semi-global block to $ \mathcal{O}(N) $. We innovatively regard each of rows and columns on the given feature map as a binary tree and complete interactions between any two nodes on the binary tree in linear time $ \mathcal{O}(2 \times (H-1) ) $ or $ \mathcal{O}(2 \times (W-1)) $. Thus total computational complexity is $ \mathcal{O}(H \times 2 \times (W-1)+W \times 2 \times (H-1)) = \mathcal{O}(4 \times W \times H-2 \times (W+H)) $ for $ W $ columns and $ H $ rows. Specifically, let us consider operations on one of the rows or columns, as shown in Fig. \ref{aggrProp}. We select one node arbitrarily in the given row or column as a root node to establish a binary tree. Except the root node and leaf nodes, every node in the binary tree has a parent node and a child node. Every node is supposed to capture contextual information of others in this binary tree. The process of capturing contextual information is split into aggregation and updating steps: 
\begin{itemize} 
\item We aggregate contextual information from leaf nodes towards the root node recursively according to
\begin{equation}\label{linearaggr}
A(I_u) =
\left\{
\begin{array}{lr}
I_u  \ \ \ \ \ \ \ \ \ \ \ \ \ \ \ \ \ \ \ \ \ \ \ \ \ \ \ \ \ \ \ u \  is\  a\  leaf \  node  & \\
I_u + \sum\limits_{P(v)=u}\Omega(u,v)A(I_v) \ \ \ \ \ \ \  others,		
\end{array} 
\right.
\end{equation}
where $ I_u $ is the input feature vector at node $ u $, $ P(v)=u $ means $ u $ is the parent node of $ v $. $ \Omega(u,v) $ is the defined weight function between $ u $ and $ v $ in (\ref{weighfunction_p}). 
\item  We update features from the root node towards leaf nodes recursively by
\begin{equation}\label{linearupdate}
U(I_u) =
\left\{
\begin{array}{lr}
A(I_u)  \ \ \ \ \ \ \ \ \ \ \ \ \ \ \ \ \ \ \ \ \ \ \ \ \ \ u \  is \  a \  root \  node  & \\
\Omega(P(u),u)U(I_{P(u)}) & \\ + (1-\Omega^2(u,P(u)))A(I_u) \ \ \ \ \ \ \ \ others,		
\end{array} 
\right.
\end{equation}
where $ A(I_u) $ is computed by (\ref{linearaggr}).
\end{itemize} 
The computational complexity of the aggregation and that of updating steps are both $ \mathcal{O} (H-1) $ for columns or $ \mathcal{O}(W-1) $ for rows. Thus the total computational complexity is $ \mathcal{O}(2 \times (H-1) ) $ or $ \mathcal{O}(2 \times (W-1)) $ for a single binary tree. 

Operations on other columns and rows are similar. We do not perform updating steps until the aggregation steps are completed for all binary trees. Thus, the aggregation and updating steps of each binary tree are independent of others, which can be parallel accelerated on GPUs. The output feature map is denoted as $ I' $. A node $ u $ is included in two binary trees ($ T_c, T_r $), as shown in Fig. \ref{aggrProp}. We add the updating results of $ u $ in $ T_c $ and $ T_r $ to $ I_u $:
\begin{equation}\label{Finalaggr}
I'_u = T_c(U(I_u)) + T_r(U(I_u)) + I_u,
\end{equation}
where $ I'_u $ is the feature vector at $ u $ in $ I' $, and $ I'_u $ has contained contextual information in horizontal and vertical directions of $ u $ in $ I $.

\begin{figure*}	\centering
	\centerline{\includegraphics[width=16.0cm]{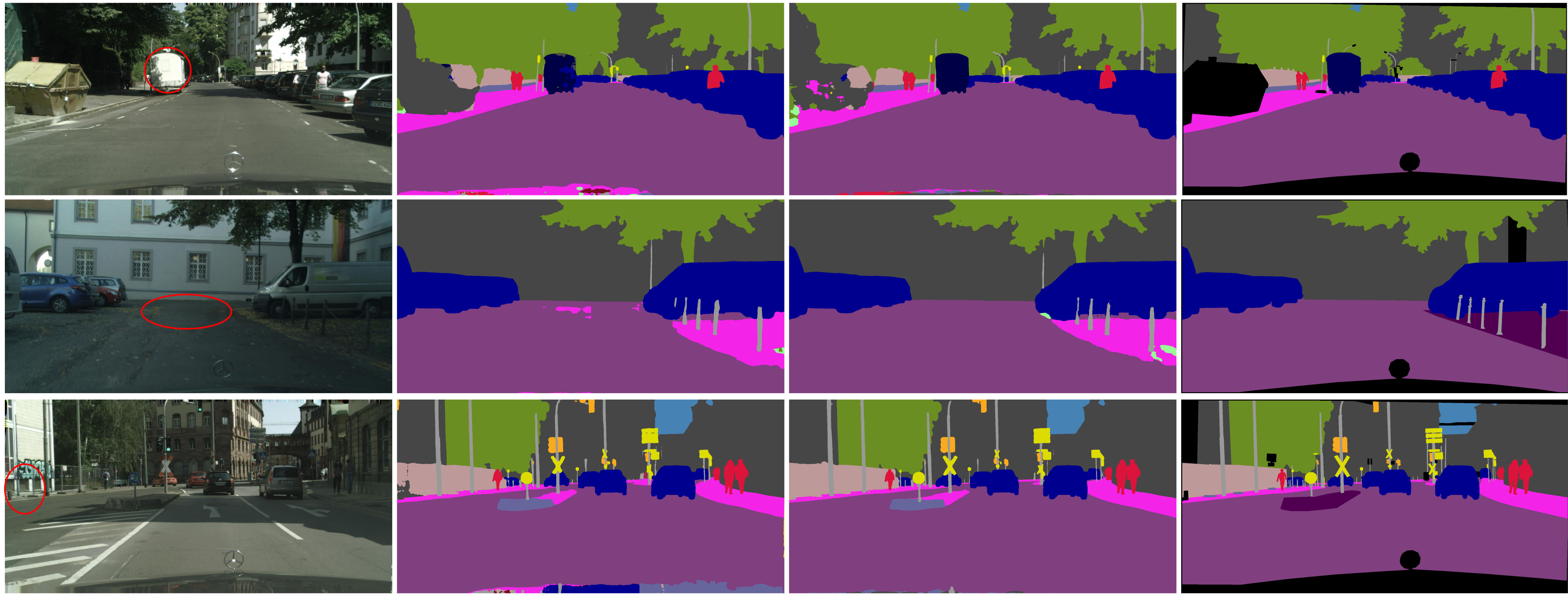}}
	\caption{Segmentation results on Cityscapes validation set. Columns from left to right are original images, segmentation results with a single semi-global block, segmentation results with hierarchical semi-global blocks and ground truths respectively. (Best viewed in color)}
	\label{segmentResults}
\end{figure*}
\section{Experiments}
In this section, we evaluate SGSNet on semantic segmentation and image retrieval. We first conduct extensive experiments on semantic segmentation to understand the behavior of SGSNet, and then extend SGSNet to image retrieval to demonstrate the generality of SGSNet.

\subsection{Experiments on Semantic Segmentation}
We adopt Cityscapes \cite{cordts2016cityscapes} benchmark for semantic segmentation and report the metrics of Mean IoU (mean of class-wise intersection over union). Cityscapes is a semantic segmentation benchmark focusing on urban street scenes. This benchmark contains 5,000 finely annotated images which are divided into 2975, 500, 1525 images as training, validation and testing set. A larger set of 20,000 coarsely annotated images are also provided for supporting methods that exploit large volumes of weakly-labeled data. The dataset also defines 30 visual classes, of which 19 classes are used in our experiments.

Based on the proposed linear time algorithm, SGSNet can be plugged into any deep neural networks conveniently for capturing long-range dependencies. We adopt ResNet-101 \cite{he2016deep} (pre-trained on ImageNet) as our backbone with minor changes. The last two down-sampling layers are dropped and dilated convolutions are embedded into subsequent convolutional layers similar to \cite{huang2019ccnet}. 

We use the mini-batch stochastic gradient descent (SGD) with a momentum of 0.9 and a weight decay of 0.0001. A poly learning rate schedule with an initial learning rate of 0.01 and power of 0.9 is employed. We also augment the training set by randomly scaling (0.75 to 2.0 $\times$) and then crop out patches with 769 $\times$ 769 pixels as the network input resolution. The learnable parameter $ \alpha $ and $\beta$ are both initialized with 1. The number of channels in $ \Lambda $ is one eighth of the number of channels in $ I $ and we share the parameters of the hierarchical semi-global blocks.

\subsubsection{Comparisons with Other Methods}
We first train SGSNet on Cityscapes training set and present results on Cityscapes validation set in Tab. \ref{table_cityscapes_val}. We can see that our method achieves better results than other methods, even if DeepLibv3+ \cite{chen2018encoder} and DPC \cite{chen2018searching} use a more powerful backbone. 

We further compare our method with existing methods on the Cityscapes testing set, as shown in Tab. \ref{table_cityscapes_test}. SGSNet is trained with only finely annotated data and then the test results are submitted to the official evaluation server. It can be seen from Tab. \ref{table_cityscapes_test} that SGSNet outperforms other methods no matter that they employ either the same backbone as ours or a stronger one \cite{yuan2018ocnet,yang2018denseaspp}. Even if CCNet freezes batch normalization (BN) layers and finetunes the model with a low learning rate after training certain iterations while we do not use any of those tricks, our method is still better than CCNet. This proves the importance of considering both feature similarity and geometric proximity for preserving object shapes when capturing long-range dependencies. Furthermore, our SGSNet is more efficient, which will be elaborated in Sec. \ref{ComputationalComplexity}. 

\begin{table}[h]
	\caption{Comparisons with other methods on Cityscapes validation set.}
	\begin{center}
		\setlength{\tabcolsep}{1.0mm}
			{
				\begin{tabular}{ c | c | c | c } 
				\hline
				Method & Backbone & multi-scale & mIoU(\%) \\
				\hline
				DeepLabv3 \cite{chen2017rethinking} & ResNet-101 & Yes & 79.3\\
				DeepLabv3+ \cite{chen2018encoder} & Xception-65 & No & 79.1\\
				DPC \cite{chen2018searching} & Xception-71 & No & 80.8\\
				CCNet \cite{huang2019ccnet}  & ResNet-101 & Yes & 81.3\\
				\hline
				SGSNet & ResNet-101 & No & 80.9\\
				SGSNet & ResNet-101 & Yes & \textbf{81.9}\\
				\hline
			    \end{tabular}
		
			}

	\end{center}
	\label{table_cityscapes_val}
\end{table}

\begin{table}[h]
	\caption{Comparisons with other methods on Cityscapes testing set, where $\divideontimes$ means this method trains with finely annotated training and validation sets. }
	\begin{center}
		\begin{tabular}{ c | c | c  } 
			\hline
			Method & Backbone  & mIoU(\%) \\
			\hline
			DeepLab-v2 \cite{chen2017deeplab} & ResNet-101  & 70.4\\
			RefineNet \cite{lin2017refinenet} $\divideontimes$ & ResNet-101 &  73.6\\
			SAC \cite{zhang2017scale} $\divideontimes$ & ResNet-101 & 78.1\\
			GCN \cite{peng2017large} $\divideontimes$ & ResNet-101  & 76.9\\
			DUC \cite{wang2018understanding} $\divideontimes$ & ResNet-101  & 77.6\\
			ResNet-38 \cite{yuan2018ocnet} & WiderResnet-38  & 78.4\\
			PSPNet \cite{zhao2017pyramid} & ResNet-101  & 78.4\\
			BiSeNet \cite{yu2018bisenet} $\divideontimes$ & ResNet-101  & 78.9\\
			AAF \cite{ke2018adaptive} & ResNet-101  & 79.1\\
			PSANet \cite{zhao2018psanet} $\divideontimes$ & ResNet-101  & 80.1\\
			DFN \cite{yu2018learning} $\divideontimes$ & ResNet-101  & 79.3\\
			DenseASPP \cite{yang2018denseaspp} $\divideontimes$ & DenseNet-161  & 80.6\\	
			TF \cite{song2019learnable} $\divideontimes$  & ResNet-101 & 80.8\\
			CCNet \cite{huang2019ccnet} $\divideontimes$  & ResNet-101 & 81.4\\
			\hline
			SGSNet  $\divideontimes$  & ResNet-101 & \textbf{82.1}\\
			\hline
		\end{tabular}
	\end{center}
	\label{table_cityscapes_test}
\end{table}
\subsubsection{Analysis of Computational Complexity}
\label{ComputationalComplexity}
Based on the linear time algorithm described in Sec. \ref{LinearTime}, the computational complexity of SGSNet is $ \mathcal{O}(N) $, which is linearly proportional to the number of nodes in a feature map. We explore the computation cost and the number of parameters of SGSNet on Cityscapes validation set. We also use ResNet-101 as the backbone and the input size is 769 $\times$ 769 pixels, thus the size of input feature maps of the hierarchical semi-global blocks is 97 $\times$ 97 pixels. The baseline network is ResNet101 with some minor changes, where dilated convolutional layers are adopted at stage 4 and 5. As shown in Tab. \ref{table_cityscapes_efficiency}, SGSNet improves the performance by 5.8\% mIoU over the baseline with additional 0.295M parameters and 11.4G FLOPs overheads. We also list the GFLOPs and parameters of Non-local and CCNet in Tab. \ref{table_cityscapes_efficiency}. It can be seen that SGSNet achieves higher performance while uses less parameters and FLOPs than CCNet. Non-local uses slightly less parameters than SGSNet, but the additional FLOPs of Non-local is far greater (108G VS 11.4G) than SGSNet. Also, SGSNet increases mIoU by 1.8\% compared with Non-local. Therefore, SGSNet is more efficient and effective than the other two.
\begin{table}[h]
	\caption{Comparisons of SGSNet with CCNet and Non-local. The increments of FLOPs and parameters are estimated for an input of 1 $\times$3$\times$769$\times$769 pixels.}
	\begin{center}
		\begin{tabular}{ c | c | c | c } 
			\hline
			Method & GFLOPs ($ \blacktriangle $) & Params (M $ \blacktriangle $) & mIoU(\%) \\
			baseline & 0 & 0 & 75.1\\
			Non-local & 108 & 0.131 & 79.1\\
			CCNet & 16.5 & 0.328 & 79.8\\
			SGSNet & 11.4 & 0.295 & 80.9\\
			\hline
		\end{tabular}
	\end{center}
	\label{table_cityscapes_efficiency}
	\vspace{-1.1em}
\end{table}

\begin{figure}
	\centering
	\centerline{\includegraphics[width=8.0cm]{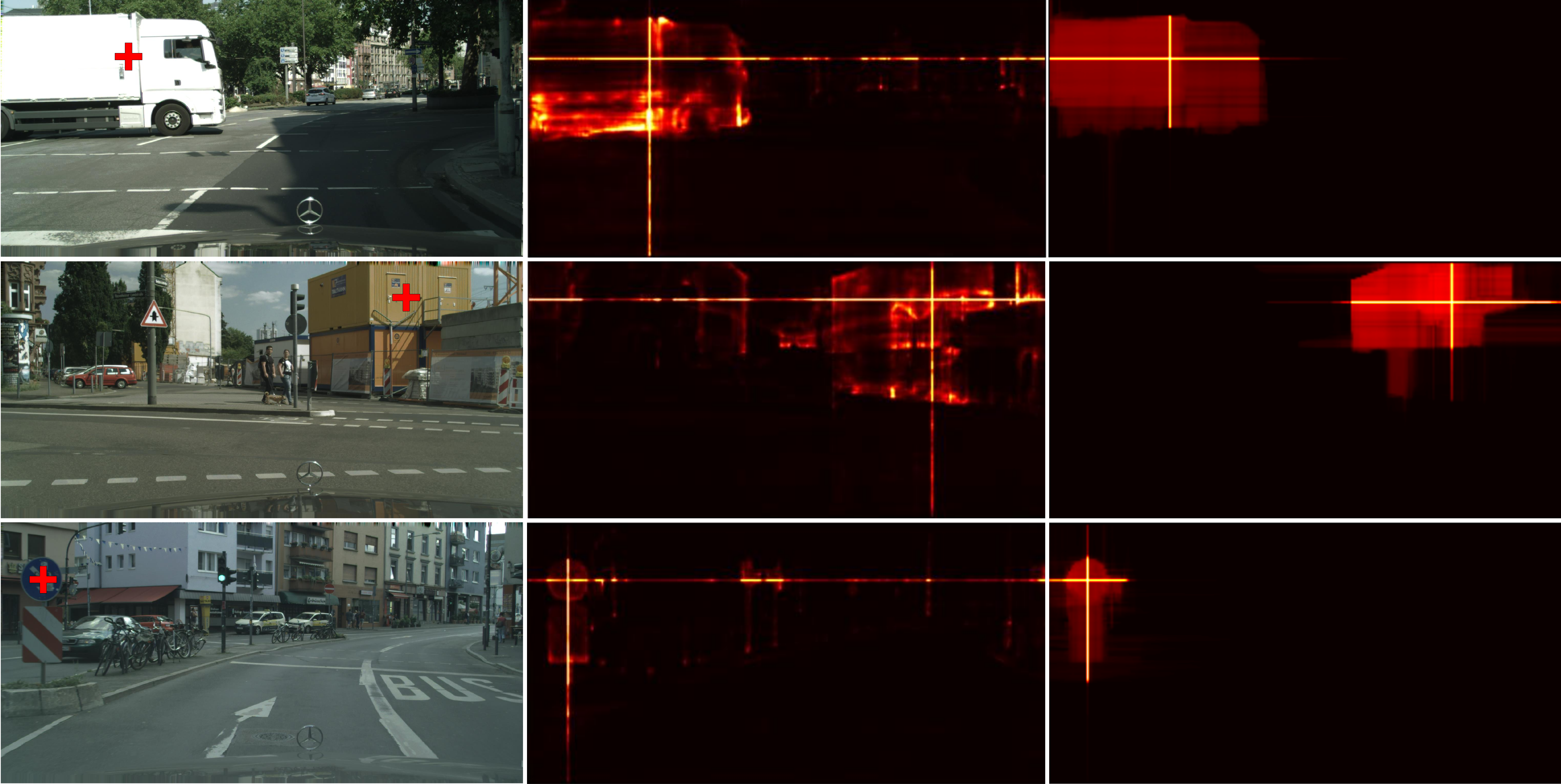}}
	\caption{Visualization of attention maps in a specific position (marked by a cross in red) of SGSNet and CCNet. The left are source images; the middle are the attention maps of CCNet; the right are the attention maps of SGSNet. (Best viewed in color)}
	\label{attention_scca}
\end{figure}
\subsubsection{Ablation Studies}
In order to further understand the behavior of SGSNet, we implement substantial ablation studies on Cityscapes validation set. The increments of FLOPs and parameters with different numbers of the hierarchical semi-global blocks are listed in Tab. \ref{table_cityscapes_hier}. Adding a semi-global block increases 4.4\% mIoU compared with the baseline network, which indicates importance of the semi-global block. We further add two semi-global blocks hierarchically and the performance is increased by another 1.4\% mIoU. 
These results demonstrate that the proposed hierarchical semi-global blocks can capture full-image contextual information and significantly improve the performance. We believe that the performance can be further improved by adding more semi-global blocks hierarchically. We also can see from Tab. \ref{table_cityscapes_hier} that a single semi-global block just increases 5.70G FLOPs and 0.295M parameters. Parameters do not increase more when adding the second block because they share parameters with each other. Qualitative results are also given in Fig. \ref{segmentResults}. Areas (truck, ground and fence) in red circles of the first column images are easily classified erroneously. Those areas can't be classified correctly by just adding one semi-global block, as shown at the second column in Fig. \ref{segmentResults}. But when we add two semi-global blocks hierarchically, those areas are rectified, as shown at the following third column, which explicitly demonstrates the advantages of hierarchical semi-global blocks.

We also explore the influence of learnable parameters $\alpha$ and $\beta$. We use two semi-global blocks and remove $\alpha$ and $\beta$ in (\ref{weighfunction_p}). The results are listed at the fourth row in Tab. \ref{table_cityscapes_hier}. As we can see, the performance is dropped by 0.7\% mIoU (compared with the last line in Tab. \ref{table_cityscapes_hier}) due to the absence of $\alpha$ and $\beta$.
\begin{table}[h]
	\caption{Performance of adding different numbers of hierarchical semi-global blocks to the baseline. The increments of FLOPs and parameters are estimated for an input of 1 $\times$3$\times$769$\times$769 pixels. }
	\begin{center}
		\setlength{\tabcolsep}{1.0mm}
			{
				\begin{tabular}{ c | c | c | c } 
				\hline
				Hierarchical & GFLOPs ($ \blacktriangle $) & Params (M $ \blacktriangle $) & mIoU(\%) \\
				baseline & 0 & 0 & 75.1\\
				H=1 & 5.70 & 0.295 & 79.5\\
				H=2(no $\alpha$$\beta$) & 11.4 & 0.295 & 80.2\\
				H=2 & 11.4 & 0.295 & 80.9\\
				
				\hline
			    \end{tabular}
		   }

	\end{center}
	\label{table_cityscapes_hier}
	\vspace{-1.1em}
\end{table}

\subsubsection{Visualization of Attention Module}
\label{visualization}
To further elaborate what SGSNet has learned, we visualize the learned attention maps of SGSNet in the last column of Fig. \ref{attention_scca}. For each image in the left column, we choose a specific position (marked by a cross in red) and show its corresponding attention map of SGSNet in the right column. Images in the middle column are the corresponding attention maps of CCNet. Both SGSNet and CCNet use two blocks. We can observe that, compared with CCNet, 
SGSNet can capture clear semantic similarity and learn object shapes simultaneously. For example, in the first image, the marked position in truck obtains almost all high responses from positions in truck and we can see the outline of the truck in the corresponding attention map of SGSNet clearly. Similar phenomena can be seen at the marked positions (in the building and traffic sign) of the second and third images. Areas in same objects are activated and have the high responses, which means that the proposed SGSNet can be aware of object shapes when capturing long-range dependencies. 

To classify a given ambiguous pixel, humans usually look around the pixel, rather than farther pixels, to look for contextual cues that help classify the given pixel correctly. From this perspective, SGSNet is more similar to human behavior than CCNet.

\subsection{Experiments on Image Retrieval}
\subsubsection{Datasets}
In this section, we investigate the performance of our SGSNet on large-scale image retrieval task. We train networks on Google Landmarks 18 (GL18) \cite{teichmann2019detect} and test on Revisited Oxford and Paris datasets \cite{radenovic2018revisiting}. GL18 dataset is based on the original Kaggle challenge dataset \cite{noh2016image}. It contains more than 1.2 million photos collected from 15,000 landmarks covering a wide-range of scenes (such as: historic cities, metropolitan areas, nature scenery, etc.). Revisited Oxford and Paris datasets are frequently-used to evaluate the performances of large-scale image retrieval methods. They improve Oxford and Paris datasets by dropping wrong annotations and adding new images, resulting in 4,993 and 6,322 images for Revisited Oxford and Paris datasets respectively. According to difficulty levels, the evaluation tasks are divided into three groups: \textit{easy}, \textit{medium}, and \textit{hard}. In each task, the metrics are mean average precision (mAP) and mean precision at rank 10.
\begin{table}[h]
	\caption{Comparisons on image retrieval, where Revisited Oxford and Paris datasets are abbreviated as ROxf and RPar respectively.}
	\begin{center}
		{
			\begin{tabular}{ c | c | c || c | c } 
				\hline
				\multirow{2}{*}{Method} & \multicolumn{2}{c||}{\textbf{Medium}} & \multicolumn{2}{c}{\textbf{Hard}} \\
				&ROxf&RPar&ROxf&RPar\\
				\hline
				baseline& 67.6 & 80.9 & 44.9 & 61.9 \\
				+Non-local& 68.8 & 82.0 & 46.6 & 64.5 \\
				+CCNet& 68.8 & 81.8 & 46.8 & 64.7 \\
				\hline
				+SGSNet& 68.8 & 82.0 & \textbf{47.1} & \textbf{64.8} \\
				\hline
			\end{tabular}
		}
	\end{center}
	\label{table_image_retrieval}
	\vspace{-1.1em}
\end{table}
\subsubsection{Comparisions with Other Methods}
In this section, we compare SGSNet with Non-local and CCNet on image retrieval. We adopt ResNet101-GeM \cite{radenovic2018fine} trained with the triplet loss and second-order similarity (SOS) loss \cite{tian2019sosnet} as our baseline. ResNet101 \cite{he2016deep} contains five fully-convolutional blocks conv1 to conv5\_x. For fair comparison, we just add Non-local, CCNet and SGSNet after conv5\_x respectively. Inside of Non-local, CCNet and SGSNet, the number of channels $ C $ of an input feature map $ I $ is reduced to $ C/8 $ for efficient computations. We also do not use batch normalization and Relu layers. Following \cite{ng2020solar}, we report mAP of these methods in medium and hard tasks of Revisited Oxford and Paris datasets, as shown in Tab. \ref{table_image_retrieval}. 

From Tab. \ref{table_image_retrieval}, we can see that adding SGSNet to the baseline can significantly improve accuracies of image retrieval task. Compared with Non-local and CCNet, SGSNet achieves comparable results in the medium task but superior results in the hard task. 
There are lots of large viewpoint changes and significant occlusions in the hard task, which will influence the aggregation of contextual information. Most of contextual information is captured around central nodes for SGSNet and thus large viewpoint changes and occlusion have relatively less impact on SGSNet. As a whole, SGSNet achieves better results. While, SGSNet takes less computation cost as shown in Sec. \ref{ComputationalComplexity}. 
 
\section{Conclusion}
In this work, we propose a semi-global shape-aware network (SGSNet) which considers both feature similarity and geometric proximity for preserving object shapes when capturing long-range dependencies. 
Each position in the feature map captures contextual information in horizontal and vertical directions according to both similarity and proximity in a single block, and then harvests entire image contextual information by adding more semi-global blocks hierarchically. In addition, each of rows and columns on the given feature map is regarded as a binary tree. Then based on structures of the binary trees, we present a linear time algorithm further improving the efficiency of SGSNet. Extensive ablation studies have been conducted to deeply understand the proposed method. We also show the superiorities of SGSNet on semantic segmentation and image retrieval. In the future, we will explore SGSNet in more vision tasks.


{\small
\bibliographystyle{ieee_fullname}
\bibliography{egbib}
}

\end{document}